\documentclass[conference]{IEEEtran}
\IEEEoverridecommandlockouts
\usepackage{cite}
\usepackage{amsmath,amssymb,amsfonts}
\usepackage{algorithmic}
\usepackage{graphicx}
\usepackage{textcomp}
\usepackage{xcolor}
\usepackage{multirow}
\usepackage{balance}

\def\BibTeX{{\rm B\kern-.05em{\sc i\kern-.025em b}\kern-.08em
    T\kern-.1667em\lower.7ex\hbox{E}\kern-.125emX}}

\begin{document}

\title{Thermal Face Image Classification using Deep Learning Techniques \\
}

\author{\IEEEauthorblockN{Prosenjit Chatterjee}
\IEEEauthorblockA{\textit{Dept. of Computer Science and Cyber Security} \\
\textit{Southern Utah University}\\
Cedar City, UT, USA \\
prosenjitchatterjee@suu.edu}
*Corresponding author
~\\
\and
\IEEEauthorblockN{ANK Zaman}
\IEEEauthorblockA{\textit{Dept. of Physics and Computer Science} \\
\textit{Wilfrid Laurier University}\\
Waterloo ON, Canada \\
azaman@wlu.ca}
~\\

}

\maketitle

\begin{abstract}
Thermal images have various applications in security, medical and industrial domains. This paper proposes a practical deep-learning approach for thermal image classification. Accurate and efficient classification of thermal images poses a significant challenge across various fields due to the complex image content and the scarcity of annotated datasets. This work uses a convolutional neural network (CNN) architecture, specifically ResNet-50 and VGGNet-19, to extract features from thermal images. This work also applied Kalman filter on thermal input images for image denoising.  The experimental results demonstrate the effectiveness of the proposed approach in terms of accuracy and efficiency.
\end{abstract}

\begin{IEEEkeywords}
Thermal face image, Deep Learning, ResNet-50, VGGNet-19, Kalman filter, Image classification
\end{IEEEkeywords}

\section{Introduction}\label{int}
Thermal imaging technology has been widely used in various fields, such as security, medical, and industrial applications. Thermal images capture the infrared radiation emitted from an object or a scene, which can provide valuable information that is not visible to the human eye. However, accurate and efficient classification of thermal images remains a challenging task due to the complexity of the image content and the lack of available annotated datasets. Convolutional neural networks (CNNs) have been widely used for feature extraction and classification in image processing. However, training a deep CNN from scratch on a small annotated dataset can lead to overfitting and poor generalization performance.
This paper proposes a novel deep-learning approach for thermal image classification. Specifically, we use a pre-trained ResNet-50 model to extract features from thermal images and fine-tune the model on a small annotated dataset. We evaluate the performance of the proposed approaches on two publicly available thermal image datasets.

The rest of the paper is organized as follows:
section~\ref{rw} talks about the literature review, Section~\ref{d} describes the dataset used in this research, Section~\ref{model} represents the technical details of the models implemented, Section~\ref{res} represents experimental results, and Section~\ref{con} describes the concluding remarks of this paper.


\section{LITERATURE REVIEW}\label{rw}
\subsection{Image Quality}
In the study of image processing, the quality of the image plays a crucial role. There are several state-of-the-art approaches utilized to improve the quality of image samples. Some of the popular approaches are based on prediction~\cite{keys1981cubic}, image edge detection~\cite{freedman2011image}, image representations~\cite{yang2010image}, etc. Image interpolation becomes popular, especially when interpolation is embedded with the nearest neighbor~\cite{rukundo2012nearest}.   

\subsection{Thermal Image Classification}
Thermal imaging, also known as infrared imaging, is a technology that captures the temperature variations in an object or scene and converts them into a visible image. In general, thermal images are typically captured through a Thermal Camera and/or Infrared Radiation camera, using thermal sensors. A thermal image is a kind of compromised image, that has fused edges in common. Therefore, thermal images can be categorized as low-resolution (LR) images. However, an accurate classification of thermal images is crucial for extracting meaningful information.
Mathur, P., et. al.~\cite{mathur2021real} (2021) experimented on LR images to obtain super-resolution (SR) images on thermal cameras through an optimized pipeline on an embedded edge device. 
Choi, Y. et. al.,~\cite{choi2016thermal} (2016) addressed the need for improved recognition under challenging conditions, such as low-light or erratic illumination. The authors propose a method called 'Thermal Image Enhancement using a Convolutional Neural Network' (TEN), which utilizes CNNs to enhance low-resolution thermal images, guided by RGB data~\cite{choi2016thermal}. They demonstrated that their approaches enhanced visibility and improved the performance of various object recognition tasks, including pedestrian detection, visual odometry, and image registration~\cite{choi2016thermal}.
The study presented by Rajinikanth, V. et. al.,~\cite{rajinikanth2021breast} (2021) focuses on the development of an automated system for breast cancer detection using Breast-Thermal-Images (BTI). The approach involves recording images from various breast orientations, extracting healthy/ductal carcinoma in situ (DCIS) image patches, applying image processing techniques, extracting features like Gray Level Co-Occurrence Matrix (GLCM) and Local Binary Pattern (LBP) with varied weights, optimizing these features using the Marine-Predators-Algorithm (MPA), and performing two- class classification. The Decision-Tree (DT) classifier is found to achieve enhanced accuracy ($>$92\%) compared to other methods. The study demonstrates that the proposed scheme works effectively on clinical-grade breast thermal images and improves disease detection accuracy.

Gao, Z. et.  al.,~\cite{gao2020extracting} (2020) explored in their study about the use of infrared cameras in dark and foggy conditions and focused on feature extraction from infrared images using deep neural networks. Traditional CNNs trained on visible light images are not suitable for infrared images. The researchers retrain the VGG-19 CNNs architecture using a thermal image dataset obtained from public sources. Data augmentation techniques such as flipping, zooming, shifting, and rotating the images are applied. Transfer learning is used to fine-tune the trainable layers of the VGG-19 network. The results show that the transfer-learned neural network can extract more information from infrared images compared to the original network. The results show that the transfer-learned neural network can extract more information from infrared images than the original network. The method is also applied to MobileNet, which yields improved results. Infrared imaging is essential in unmanned vehicles and target detection applications. While deep CNNs are widely used for feature extraction in visible images, this study is one of the first to explore feature extraction from infrared images using CNNs and transfer learning.

The paper by Adam Glowacz~\cite{glowacz2021fault} (2021) discusses a method for fault diagnosis in electric impact drills using thermal imaging. The author introduces a novel technique called BCAoID (Binarized Common Areas of Image Differences) for feature extraction from thermal images. The study analyzes thermal images of three electric impact drills: a healthy one, one with a faulty fan (with ten broken fan blades), and one with a damaged gear train. The extracted features are then analyzed using the Nearest Neighbor classifier and a backpropagation neural network. The fault recognition results range from 97.91\% to 100\%, demonstrating the effectiveness of this approach for diagnosing faults in rotating machinery and engines. The paper highlights the potential economic benefits of using thermal imaging for maintenance and safety in various industrial applications, including power tools.

\subsection{Deep Learning}
Deep learning (DL) a popular subset of Machine Learning (ML) transforms significantly during the past decades. Many researchers relentlessly achieved revolutionary research in deep learning including three Turing Award winners in Deep Learning from the year 2015 to 2020. Among these pioneering works, Krizhevsky, A. et. al.,~\cite{krizhevsky2017imagenet} (2012) marked a pivotal moment in first introducing ImageNet classification with deep convolutional neural networks (CNNs). The groundbreaking research~\cite{choi2016thermal} demonstrated the potential of deep neural networks to automatically learn hierarchical features from data, inspiring a wave of innovations in computer vision. The impact of "ImageNet Classification"~\cite{krizhevsky2017imagenet} extended beyond computer vision. Its architectural advancements, including the use of rectified linear units (ReLUs) and deep convolutional layers, provided a blueprint for subsequent deep learning models~\cite{krizhevsky2017imagenet}. These innovations laid the foundation for architectures like VGG, ResNet, and Inception, pushing the boundaries of image recognition accuracy.
Beyond image classification, deep learn- ing impacted reinforcement learning, as showcased by Mnih, V. et. al.,~\cite{mnih2013playing} (2013) in their significant contribution to reinforcement learning. While originally framed as a reinforcement learning~\cite{mnih2013playing}, this work demonstrated the power of deep Q-learning in learning complex visual representations from raw pixel inputs. Deep Q-learning is a reinforcement learning algorithm that uses a deep neural network and approximates the Q-function, which is used to determine the optimal action to take in each state. The successful application of deep reinforcement learning to Atari 2600 games was a testament to the adaptability of deep learning models in diverse computer vision scenarios.
Furthermore, Goodfellow, I., et al.,~\cite{goodfellow2014generative} (2014) introduced Generative Adversarial Networks (GANs), which unlocked new possibilities in image generation and style transfer. GANs [12], with their generator-discriminator architecture, have produced realistic synthetic image samples and found applications in image-to-image translation and data augmentation, further enriching the computer vision toolbox.
Addressing the challenges of training deep neural networks, "Deep Residual Learning for Image Recognition" by Kaiming He et al.,~\cite{he2016deep} (2016) brought the ResNet architecture. ResNet's ingenious use of skip connections mitigated the vanishing gradient problem, enabling the training of exceptionally deep networks. The deep learning networks exhibited superior image recognition accuracy and forever changed the landscape of deep learning in computer vision.
As deep learning continues to evolve, these foundational papers remain a testament to the enduring impact of neural networks and their profound influence on artificial intelligence.

\section{Dataset}\label{d}
Tufts~\cite{tufts2020} is one of the most comprehensive, large-scale, multi-modality face datasets freely available to the public. This initiative helps other researchers in developing face recognition methods across multimodal images. By incorporating photograph images, thermal images, near-infrared images, recorded video, computerized facial sketches, and 3D images, the team has accumulated a collection of more than 10,000 images from 113 individual volunteers. The volunteers represent diverse ethnicities, genders, and countries of origin. For this research, the collection of thermal images is used for experimentation. For training, 978 images belong to 109 classes, and 457 images belong to 51 classes for testing.

Charlotte-ThermalFace~\cite{Charlotte, ASHRAFI2022104209} is a thermal face dataset with the facial landmarks, ambient temperature, relative humidity, the air speed of the room, distance to the camera, and the subject thermal sensation at the time of capturing images. It contains approximately 10,000 infrared thermal images from 10 subjects in varying thermal conditions, at several distances from the camera, and various head positions. It is the first public facial thermal dataset fully annotated with the environmental properties, including air temperature, relative humidity, airspeed, distance from the camera, and subjective thermal sensation of each person at the time. There are 72 or 43 landmarks manually annotated for every dataset image. For this research, to train a model, 10376 images with 10 classes, and for testing, 1306 images with 5 classes are used.

\subsection{VGG Net}
The Visual Geometry Group (VGG) Networks have been vital in shaping deep learning and computer vision models. The VGG architecture, proposed by Simonyan, K., et. al.,~\cite{simonyan2014very} (2014) introduced a straightforward yet powerful design that significantly improved the performance of image classification. VGG Nets marked a notable shift in the depth of convolutional neural networks (CNNs), offering key insights into network design and optimization.
VGG Nets, particularly the VGG16 and VGG19 variants, are characterized by their depth and uniform architecture [14]. They consist of a series of convolutional layers, each followed by max-pooling layers, and at the end, connected through fully connected layers for classification~\cite{krizhevsky2017imagenet}. The key innovation in VGG Nets was using of small 3x3 convolutional filters stacked multiple times, which allowed the network to learn hierarchical features effectively~\cite{simonyan2014very}.
Szegedy et al.,~\cite{szegedy2015going} (2015) presented the Inception architecture, demonstrating that the depth of VGG could be further increased while maintaining computational efficiency. The Inception model employed a combination of convolutional filters of varying sizes within a single layer, reducing the number of parameters and computational cost compared to VGG Nets.
Zeiler, M. D. et. al.,~\cite{zeiler2014visualizing} (2013) experimented with how the inner layers work in CNNs and their importance in VGG Nets. In another experiment, Szegedy, C. et. al.,~\cite{szegedy2016rethinking} proposed a fused experiment on Inception and ResNets.

%

\subsection{ResNet}
He, K. et. al.,~\cite{zeiler2014visualizing} (2016) proposed and demonstrated Residual Networks (ResNets), a phenomenal deep learning architecture that has a high impact on the computer vision field of research. The authors~\cite{he2016deep} addressed a significant challenge in training very deep neural networks: the vanishing gradient problem, by introducing residual connections, or skip connections between layers. Residual connections fundamentally changed the way deep networks were designed. Instead of hoping each layer would learn the ideal transformation, residual networks allowed layers to learn residual functions, which are then added to the input [13]. This architecture allowed for the training of extremely deep networks, effectively mitigating the vanishing gradient issue~\cite{zeiler2014visualizing}.
Later, in another research experiment, He, K. et.al.,~\cite{he2016deep} (2016) investigated the importance of identity mapping, enabling very deep network training and model performance. 

Zagoruyko, S. et. al.,~\cite{zagoruyko2016wide} (2016) proposed and implemented a wide ResNet, which increased the width of ResNets to achieve better accuracy with less depth showcasing the adaptability of the ResNet concept.
Xie, S. et. al.,~\cite{xie2017aggregated} (2017) introduced ResNetXt, an aggregated residual transformation for deep neural networks. The proposed and implemented architecture utilized grouped convolutions to capture richer feature representations and achieved state-of-the-art results on various image recognition tasks, illustrating the potential for customization within the ResNet framework.
In a research article, Shah A. et. al.,~\cite{shah2016deep} (2016) exploited exponential linear unit (ELU) in combination with ResNet to establish the positive impact in improving training convergence and classification accuracy, offering valuable insights for model optimization.
Residual Attention Networks (RANs) marked another significant advancement. In an experiment by Wang, Fei. et. al.,~\cite{wang2017residual} on RANs on image classification. RANs incorporated attention mechanisms within residual blocks, enabling them to focus on informative regions of an image~\cite{wang2017residual}. RANs excelled in tasks requiring fine-grained discrimination, showcasing the versatility of the ResNet architecture~\cite{wang2017residual}.

\section{MODEL DESCRIPTION}\label{model}

\subsection{Image Pre-Processing }
\begin{figure}[!ht]
	\includegraphics[scale=.35]{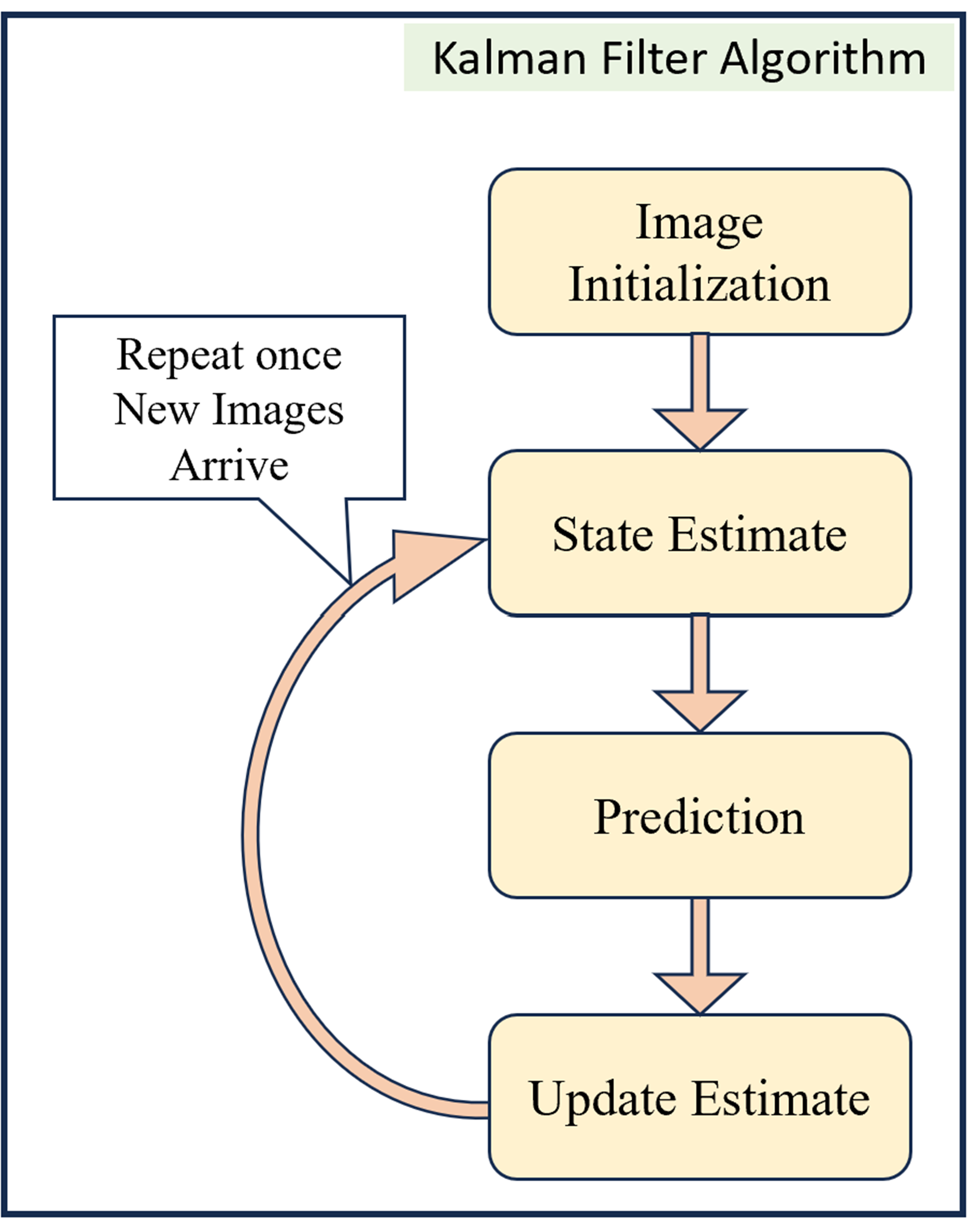}
	\centering
	\caption{Kalman Filter Algorithm}
	\label{fig:kalman}
\end{figure}

We obtained thermal image data from Charlotte and Tufts Thermal Face datasets. As we know, infrared-enabled thermal images are low-resolution images. Also, it is known that thermal image samples are full of noise. To reduce the external environment and background noise, we planned to implement a Kalman filter on individual image samples as a preprocessing step before classification through deep learning. Figure-~\ref{fig:kalman} shows below the steps on how Kalman Filter works for noisy image filtering. 

Kalman filter algorithm has the following steps:

\renewcommand{\labelenumi}{\alph{enumi})}
\begin{enumerate}
	\item Initialization: This step sets up the Kalman filter with an initial estimate of the true image and its uncertainty.
	\item State Estimate: This represents the current estimate of the true image, aiming to recover from the noisy observation.
	\item Prediction: Step to estimate how the true image is expected to change over time based on the previous state estimate and the system dynamics. This step involves applying a filter or transformation.
	\item Update: Compare the predicted state with the observed noisy image to calculate the Kalman gain. Thereafter, update the state estimate based on this gain and the difference between the prediction and observation.
	\item Repeat: Continue the prediction and update steps as new noisy images are received, iteratively refining the estimate of the true image.
\end{enumerate}

\subsection{VGGNets-19}
\begin{figure}[!ht]
	\includegraphics[scale=.48]{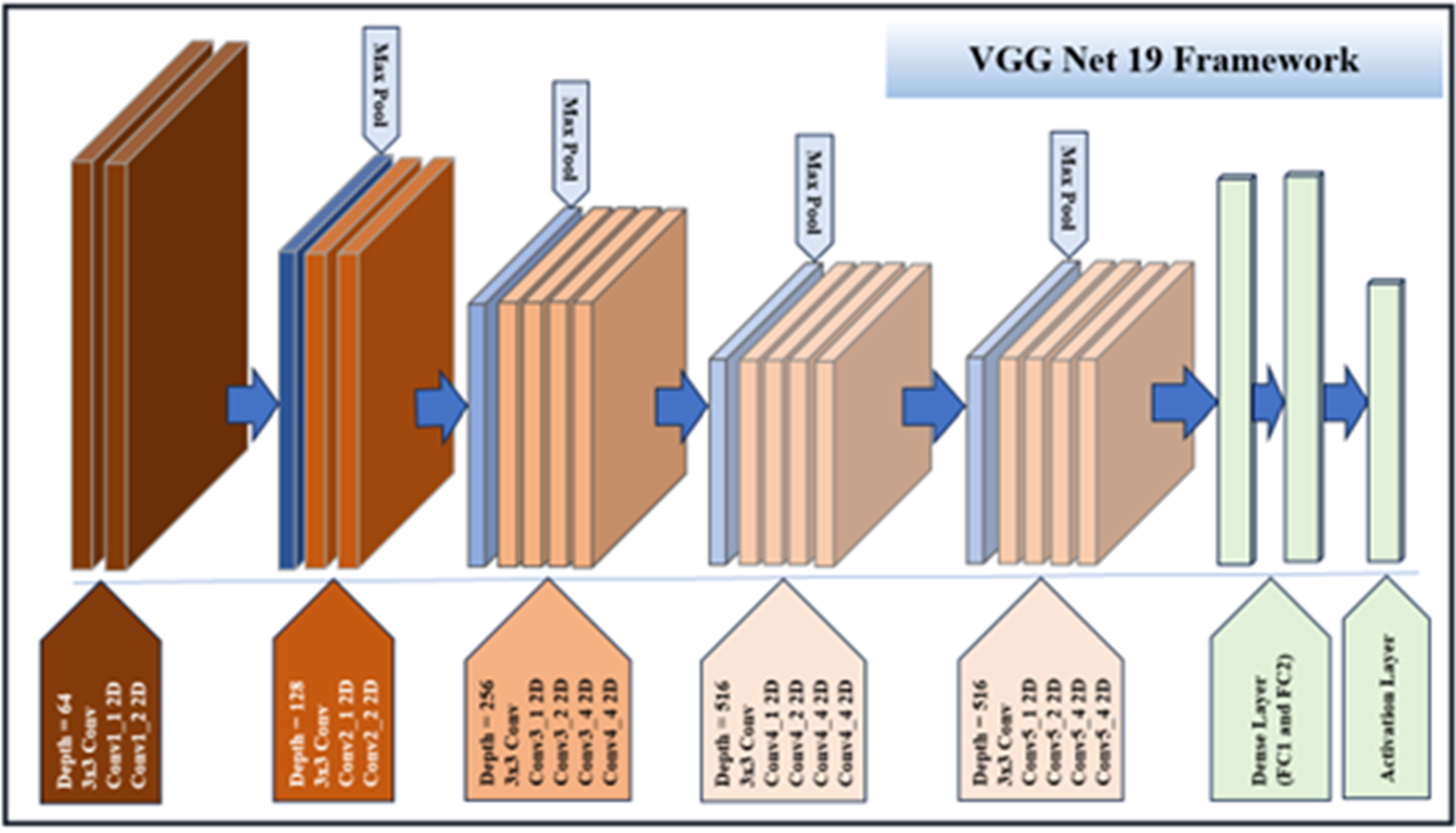}
	\centering
	\caption{VGGNet-19 Architecture}
	\label{fig:vgg}
\end{figure}

Our implemented VGGNet consists of 19 Convolutional 2D layers including 3 max pool layers, 2 fully connected layers, and 1 dense layer. Figure-~\ref{fig:vgg} shows the implemented VGGNet 19 frame structure and Table-~\ref{table:vgg19t} shows the framework with layers distribution and final flattened layers, dense layers structure, and their parameter handling capabilities.

\begin{table}[!ht]
	\centering
	\caption{VGGNet-19 Framework}
	\label{table:vgg19t}
	\resizebox{\columnwidth}{!}{%
		\begin{tabular}{|c|ccc|}
			\hline
			\multirow{2}{*}{Sr. No.} & \multicolumn{3}{c|}{VGG 19 Frame in Details}                                             \\ \cline{2-4} 
			& \multicolumn{1}{c|}{Layer (Type)}        & \multicolumn{1}{c|}{Output Shape} & Parameter \\ \hline
			1 & \multicolumn{1}{c|}{VGG19   (Functional)} & \multicolumn{1}{c|}{(None, 4, 4, 512)} & 20024384 \\ \hline
			2                        & \multicolumn{1}{c|}{flatten   (Flatten)} & \multicolumn{1}{c|}{(None, 8192)} & 0         \\ \hline
			3                        & \multicolumn{1}{c|}{dense   (Dense)}     & \multicolumn{1}{c|}{(None, 256)}  & 2097408   \\ \hline
			4                        & \multicolumn{1}{c|}{dropout   (Dropout)} & \multicolumn{1}{c|}{(None, 256)}  & 0         \\ \hline
			5                        & \multicolumn{1}{c|}{dense   (Dense)}     & \multicolumn{1}{c|}{(None, 1)}    & 257       \\ \hline
		\end{tabular}%
}
\end{table}

\subsection{ResNets-50}
\begin{figure}[!ht]
	\includegraphics[scale=.53]{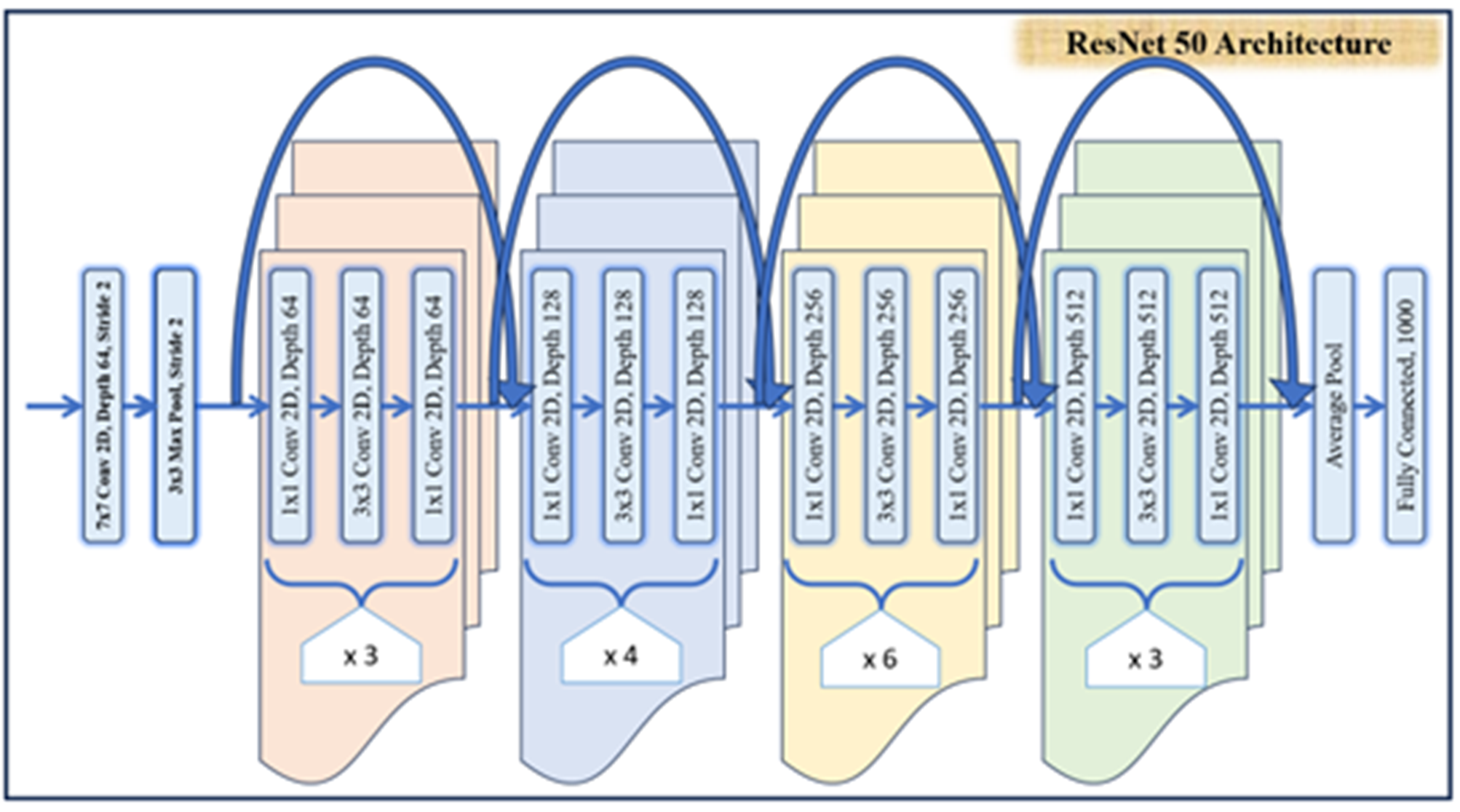}
	\centering
	\caption{ResNet-50 Framework}
	\label{fig:resnet}
\end{figure}

Our implemented ResNet-50 consists of 50 Conv 2D layers. ResNet 50 follows the principle of skip-sequential layer structure. Figure~\ref{fig:resnet} shows the implemented ResNet 50 architecture and Table-~\ref{tab:res50} shows the framework with layers distribution and the flatten layers in the final stage, dense layers structure, and their parameter handling capabilities. ResNet-50 undoubtedly handle way more parameter than VGGNet-19.

\begin{table}[]
	\centering
	\caption{ResNet-50 Framework}
	\label{tab:res50}
	\resizebox{\columnwidth}{!}{%
		\begin{tabular}{|c|ccc|}
			\hline
			\multirow{2}{*}{Sr. No.} & \multicolumn{3}{c|}{Res Net 50 Frame in Details}                                          \\ \cline{2-4} 
			& \multicolumn{1}{c|}{Layer (Type)}        & \multicolumn{1}{c|}{Output Shape}  & Parameter \\ \hline
			1 & \multicolumn{1}{c|}{ResNet   (Functional)} & \multicolumn{1}{c|}{(None, 4, 4, 2048)} & 23587712 \\ \hline
			2                        & \multicolumn{1}{c|}{flatten   (Flatten)} & \multicolumn{1}{c|}{(None, 32768)} & 0         \\ \hline
			3                        & \multicolumn{1}{c|}{dense   (Dense)}     & \multicolumn{1}{c|}{(None, 256)}   & 8388864   \\ \hline
			4                        & \multicolumn{1}{c|}{dropout   (Dropout)} & \multicolumn{1}{c|}{(None, 256)}   & 0         \\ \hline
			5                        & \multicolumn{1}{c|}{dense   (Dense)}     & \multicolumn{1}{c|}{(None, 1)}     & 257       \\ \hline
		\end{tabular}%
	}
\end{table}

\section{EXPERIMENTAL RESULTS}\label{res}

Our research experiment was conducted on a GPU cloud server environment. For the Charlotte Thermal Face datasets, we used 10376 thermal image samples under 10 classes as a training set, and 1306 image samples under 5 classes as a testing set. For the Tufts Thermal Face dataset, we utilized 978 images under 109 classes as a training set, and 457 images under 51 classes as a test set. Our implemented VGGNet-19 and ResNet-50 showed very high classification accuracy once we pre-processed the datasets through the Kalman filter denoising technique. 
\begin{table}[]
	\centering
	\caption{Experimental Results}
	\label{tab:res}
	\resizebox{\columnwidth}{!}{%
		\begin{tabular}{|c|cc|cc|}
			\hline
			\multirow{2}{*}{\begin{tabular}[c]{@{}c@{}}Classification\\  Matrices\end{tabular}} & \multicolumn{2}{c|}{CNNs: VGG Net 19}  & \multicolumn{2}{c|}{CNNs: ResNet 50}   \\ \cline{2-5} 
			&
			\multicolumn{1}{c|}{\begin{tabular}[c]{@{}c@{}}Charlotte\\  Thermal\\  Face\\  Dataset\end{tabular}} &
			\begin{tabular}[c]{@{}c@{}}Tufts\\  Thermal\\  Face\\  Dataset\end{tabular} &
			\multicolumn{1}{c|}{\begin{tabular}[c]{@{}c@{}}Charlotte\\  Thermal\\  Face\\  Dataset\end{tabular}} &
			\begin{tabular}[c]{@{}c@{}}Tufts \\ Thermal\\  Face \\ Dataset\end{tabular} \\ \hline
			Precision                                                                           & \multicolumn{1}{c|}{1}       & 1       & \multicolumn{1}{c|}{1}       & 1       \\ \hline
			Recall                                                                              & \multicolumn{1}{c|}{0.91}    & 0.9     & \multicolumn{1}{c|}{1}       & 0.9     \\ \hline
			F1 Score                                                                            & \multicolumn{1}{c|}{0.9}     & 0.99    & \multicolumn{1}{c|}{0.9}     & 0.9     \\ \hline
			Support                                                                             & \multicolumn{1}{c|}{1306}    & 457     & \multicolumn{1}{c|}{1306}    & 457     \\ \hline
			\begin{tabular}[c]{@{}c@{}}Test\\  Accuracy\end{tabular}                            & \multicolumn{1}{c|}{94.26\%} & 98.03\% & \multicolumn{1}{c|}{95.00\%} & 98.00\% \\ \hline
			\begin{tabular}[c]{@{}c@{}}Training\\  Accuracy\end{tabular}                        & \multicolumn{1}{c|}{92.26\%} & 99.08\% & \multicolumn{1}{c|}{93.27\%} & 99.10\% \\ \hline
		\end{tabular}%
	}
\end{table}

\begin{figure}[!ht]
	\includegraphics[scale=.48]{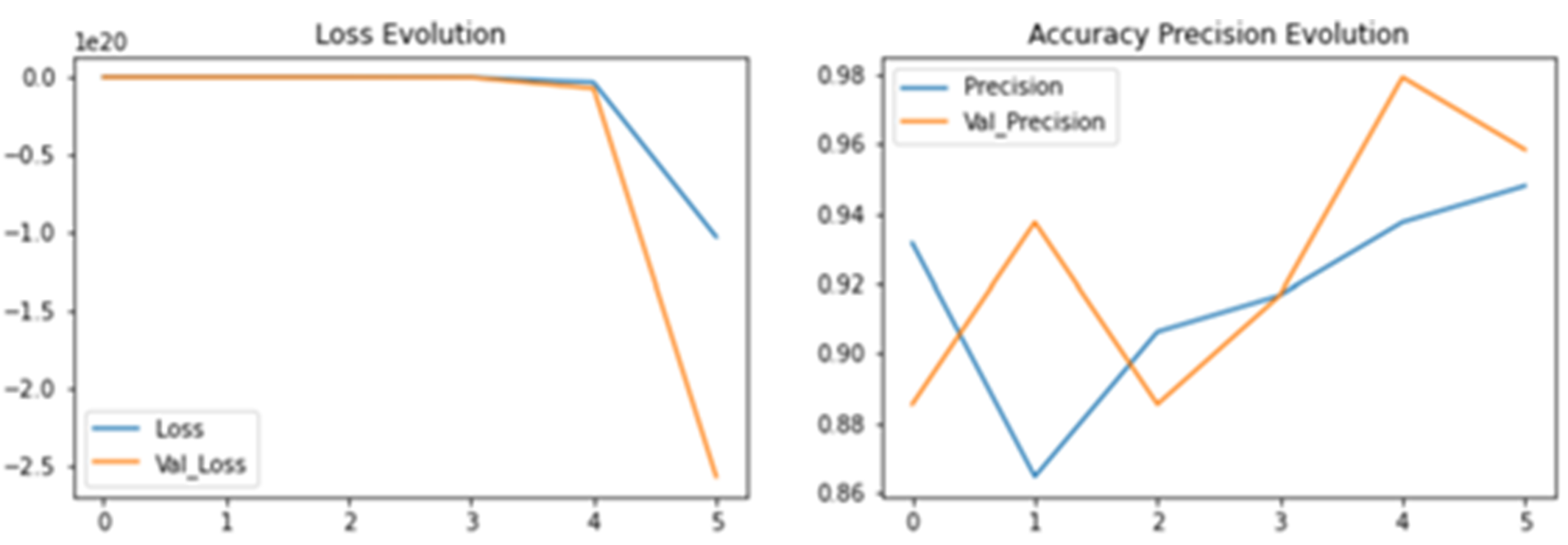}
	\centering
	\caption{Loss vs Accuracy on Charlotte Thermal Face dataset on VGGNet-19 (x-axis: loss, y-axis: Number of Epochs)}
	\label{fig:lavggc}
\end{figure}

\begin{figure}[!ht]
	\includegraphics[scale=.48]{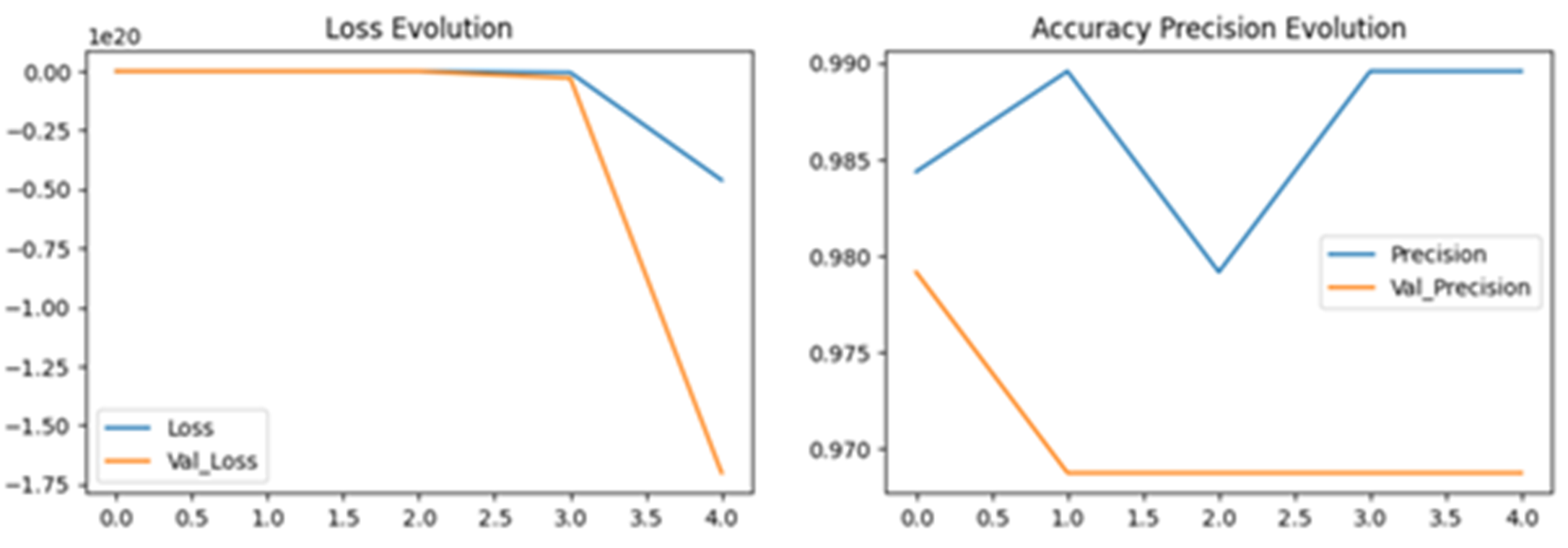}
	\centering
	\caption{Loss vs Accuracy on Tufts Thermal Face dataset on VGGNet-19 (x-axis: loss, y-axis: Number of Epochs)}
	\label{fig:lavggt}
\end{figure}

\begin{figure}[!ht]
	\includegraphics[scale=.48]{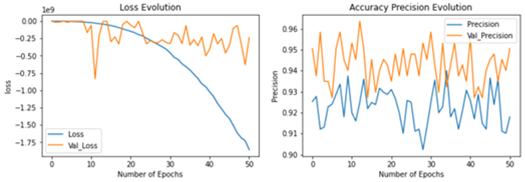}
	\centering
	\caption{Loss vs Accuracy on Charlotte Thermal Face dataset on ResNet-50 (x-axis: loss, y-axis: Number of Epochs)}
	\label{fig:laresc}
\end{figure}

\begin{figure}[!ht]
	\includegraphics[scale=.48]{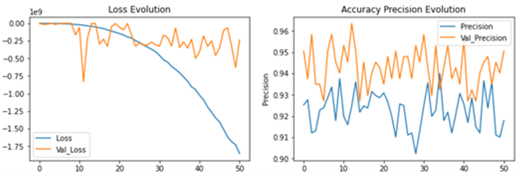}
	\centering
	\caption{Loss vs Accuracy on Tufts Thermal Face dataset on ResNet-50 (x-axis: loss, y-axis: Number of Epochs)}
	\label{fig:larest}
\end{figure}

Table-~\ref{tab:res} shows below the experimental results of VGG Nets 19 and ResNet 50 on the two different thermal face datasets. For the Tufts Thermal Face dataset, we achieved normalized training accuracy of 99\% (in the range of 99.08\% – 99.10\%), and a normalized test accuracy of 98\% from VGGNet-19 and ResNet-50 respectively. For the Charlotte Thermal Face dataset, we achieved training accuracy in the range of 92.26\% for VGGNet-19 and 93.26\% for ResNet-50. However, the test accuracy on the same dataset is 94.26\% and 95.00\% respectively for VGGNet 19 and ResNet 50. Figures-~\ref{fig:lavggc}, \ref{fig:lavggt}, \ref{fig:laresc}, and \ref{fig:larest} are presenting those results graphically.

\section{Conclusion}\label{con}
In conclusion, this paper presented a practical deep-learning approach for thermal image classification, addressing the challenges posed by the complexity of thermal image content and the limited availability of annotated datasets. The proposed approach utilized pre-trained CNNs, specifically ResNet-50 and VGGNet-19, to extract features from thermal images, then employed transfer learning to fine-tune the models on a small annotated dataset. 

The experimental results demonstrated the proposed approach's effectiveness in terms of accuracy and efficiency. Using a Kalman filter for preprocessing effectively reduced noise and enhanced the quality of the thermal images, leading to impressive classification performance. 

The implemented VGGNet-19 and ResNet-50 achieved remarkable normalized training and test accuracies on the Tufts Thermal Face dataset, indicating their ability to generalize well. Although the training accuracies were slightly lower for the Charlotte Thermal Face dataset, both models still exhibited strong performance on the test set.

The experimental results suggest that the proposed deep-learning approach and the preprocessing step can be valuable tools for various applications, including security, medical, and industrial domains, where accurate thermal image classification is crucial. Further research could explore extending this approach to more diverse datasets and real-world scenarios, potentially opening doors to even broader applications of thermal imaging technology.

\balance
\bibliographystyle{IEEEtran}
\bibliography{pz.bib}

\end{document}